\documentclass{article}

\usepackage{microtype}
\usepackage{graphicx}
\usepackage{subcaption}
\usepackage{booktabs} 
\usepackage{multirow}
\usepackage{amssymb}

\usepackage{hyperref}

\newcommand{\tablestyle}[2]{\setlength{\tabcolsep}{#1}\renewcommand{\arraystretch}{#2}\centering\footnotesize}

\usepackage[preprint]{icml2026}

\usepackage{amsmath}
\usepackage{amssymb}
\usepackage{mathtools}
\usepackage{amsthm}
\usepackage{xcolor} 
\usepackage[table]{xcolor} 
\definecolor{myorange}{RGB}{250, 235, 215}
\definecolor{mygreen}{RGB}{220, 230, 225}
\newcommand{\up}[1]{%
  \raisebox{-0.3ex}{\scriptsize\textcolor{teal}{+#1}}%
}
\newcommand{\down}[1]{%
  \raisebox{-0.3ex}{\scriptsize\textcolor{red}{-#1}}%
}
\usepackage{tabularx}
\newcolumntype{Y}{>{\centering\arraybackslash}X}

\newcommand{\gr}{\rowcolor[gray]{.96}} 

\usepackage[capitalize,noabbrev]{cleveref}

\theoremstyle{plain}

\theoremstyle{definition}

\theoremstyle{remark}

\usepackage[textsize=tiny]{todonotes}

\icmltitlerunning{Submission and Formatting Instructions for ICML 2026}

\begin{document}

\twocolumn[
  \icmltitle{ELITE: Experiential Learning and Intent-Aware Transfer\\ for Self-improving Embodied Agents}



  \icmlsetsymbol{equal}{*}

  \begin{icmlauthorlist}
    \icmlauthor{Bingqing Wei}{pku}
    \icmlauthor{Zhongyu Xia}{pku}
    \icmlauthor{Dingai Liu}{pku,intern}
    \icmlauthor{Xiaoyu Zhou}{pku}
    \icmlauthor{Zhiwei Lin}{pku}
    \icmlauthor{Yongtao Wang}{pku}
  \end{icmlauthorlist}

  \icmlaffiliation{pku}{Wangxuan Institute of Computer Technology, Peking University, Beijing, China}
  \icmlaffiliation{intern}{Completed this work during his internship at the Wangxuan Institute of Computer Technology}

  \icmlcorrespondingauthor{Yongtao Wang}{wyt@pku.edu.cn}

  \icmlkeywords{Embodied AI, Agent}

  \vskip 0.3in
]



\printAffiliationsAndNotice{}  

\begin{abstract}

Vision-language models (VLMs) have shown remarkable general capabilities, yet embodied agents built on them fail at complex tasks, often skipping critical steps, proposing invalid actions, and repeating mistakes. 
These failures arise from a fundamental gap between the static training data of VLMs and the physical interaction for embodied tasks.
%
%
To address this issue, we introduce ELITE, an embodied agent framework with {E}xperiential {L}earning and {I}ntent-aware {T}ransfer that enables agents to continuously learn from their own environment interaction experiences, and transfer acquired knowledge to procedurally similar tasks. 
ELITE operates through two synergistic mechanisms, \textit{i.e.,} self-reflective knowledge construction and intent-aware retrieval. 
Specifically, self-reflective knowledge construction extracts reusable strategies from execution trajectories and maintains an evolving strategy pool through structured refinement operations.
Then, intent-aware retrieval identifies relevant strategies from the pool and applies them to current tasks. 
Experiments on the EB-ALFRED and EB-Habitat benchmarks show that ELITE achieves 9\% and 5\% performance improvement over base VLMs in the online setting without any supervision. 
In the supervised setting, ELITE generalizes effectively to unseen task categories, achieving better performance compared to state-of-the-art training-based methods. 
These results demonstrate the effectiveness of ELITE for bridging the gap between semantic understanding and reliable action execution.

\end{abstract}

\section{Introduction}

Embodied agents operating in physical environments must decompose high-level goals into executable action sequences, adapt to environmental dynamics, and recover from errors~\cite{driess2023palme, ahn2022saycan}. While vision-language models (VLMs) have demonstrated impressive general reasoning capabilities across diverse domains~\cite{bai2025qwen25vl,zhu2025internvl3}, a fundamental gap remains between understanding the physical world and acting effectively within it~\cite{liang2023code, li2024embodiedinterface}. This gap manifests when agents skip critical intermediate steps (\textit{e.g.,} attempting to place an object without first grasping it) or propose physically invalid actions (\textit{e.g.,} opening an already-open container)~\cite{sharma2021relational, khodeir2023learning}. 
%
As these errors accumulate over time, a single misstep early in execution can invalidate subsequent actions, making task completion impossible~\cite{ross2011reduction, florence2021implicit, tu2022sample}.

These failures reflect a fundamental mismatch between how VLMs acquire knowledge and what embodied task execution demands~\cite{paolo2024position, gupta2021embodied}. 
%
VLMs are trained on static vision-language data, acquiring rich semantic understanding including object recognition, spatial reasoning, and commonsense knowledge, but lack the physical interaction with the real world~\cite{driess2023palme, ahn2022saycan}.
%
This passive observation hinders VLMs from learning action preconditions, execution ordering constraints, and error recovery strategies, which are key for embodied agents~\cite{sharma2021relational, lagrassa2022learning, khodeir2023learning}.
Humans, by contrast, learn such knowledge through embodied interaction~\cite{smith2005development}. For instance, we discover that a door must be opened before passing through by attempting to do so, not by reading about doors~\cite{vijayaraghavan2025development}. 
This experiential learning allows humans to accumulate procedural knowledge and recognize recurring patterns across tasks~\cite{gupta2021embodied, paolo2024position}. However, current embodied agents lack analogous mechanisms for learning from their own interactions.

To address this issue, previous training-based approaches adapt foundation models through behavioral cloning on expert demonstrations~\cite{kim2024openvla, zitkovich2023rt2, liu2024embodied} or reinforcement learning with shaped rewards~\cite{zhai2024rl4vlm, chen2025era, wang2025vagen}.
However, these works require substantial supervision and yield static policies that cannot incorporate new experiences after training.
Other methods augment VLM inputs with structured representations, such as scene graphs~\cite{rana2023sayplan, huang2025esca} or executable code~\cite{liang2023code}.
Though these methods provide more useful context, they lack mechanisms to update it based on execution outcomes. 
A key challenge that remains underexplored is how agents can automatically extract reusable knowledge from their own experiences and determine when such knowledge is applicable to new tasks.

This paper introduces ELITE, a framework with \textbf{E}xperiential \textbf{L}earning and \textbf{I}ntent-aware \textbf{T}ransfer for \textbf{E}mbodied Agents. The proposed framework enables embodied agents to continuously accumulate and leverage experiential knowledge. 
Specifically, ELITE comprises two components, {self-reflective knowledge construction} and {intent-aware retrieval}. 
%
First, {self-reflective knowledge construction} extracts reusable strategies from execution trajectories. We design a Reflective Experience Distiller to analyze the execution trajectory and identify effective procedures and error patterns. Then, a Context Consolidator is introduced to integrate the reflection results into an evolving strategy pool.
%
For {intent-aware retrieval}, it identifies relevant strategies for new tasks based on chain-of-thought embedding similarity, capturing procedural relationships between tasks such as ``heat the potato'' and ``warm the soup'' that share execution patterns despite differing lexical forms.

We evaluate ELITE on the EB-ALFRED and EB-Habitat benchmarks, which test embodied agents on both standard and long-horizon tasks. In the online setting without supervision, our method achieves 61\% average success on EB-ALFRED (+9 over the base VLM) and 67\% on EB-Habitat (+5). Notably, these results are obtained through fully online learning during inference, without ground-truth trajectories or task-specific supervision. In the supervised setting, ELITE generalizes effectively to unseen tasks, achieving higher performance compared to state-of-the-art training-based methods.

Our contributions are as follows:
\begin{itemize}
    \item We propose ELITE, an embodied agent framework that can accumulate procedural knowledge through self-reflective analysis of execution trajectories.
    \item We introduce an intent-aware retrieval mechanism based on chain-of-thought embedding similarity, enabling effective knowledge transfer between procedurally related tasks beyond surface-level matching.
    \item Experiments on EB-ALFRED and EB-Habitat benchmarks show that ELITE achieves state-of-the-art performance in both unsupervised online learning and supervised settings with strong generalization to unseen task categories.
\end{itemize}

\section{Related Work}

\paragraph{Foundation Model-based Embodied Agents.}
LLMs and VLMs offer a promising foundation for embodied agents due to their broad world knowledge and reasoning capabilities. Early approaches designed prompts that decompose high-level goals into executable action sequences~\cite{singh2022progprompt, song2023llmplanner, huang2022language}. Subsequent work augmented model inputs with structured representations—executable code snippets~\cite{liang2023code, silver2024generalized} or spatial abstractions such as scene graphs~\cite{rana2023sayplan, gu2024conceptgraphs}. ESCA~\cite{huang2025esca} improves spatial reasoning by injecting scene graph information into VLM-based agents.
%
However, the model knowledge in these methods is fixed, and the agents cannot update their understanding based on what they encounter during execution. 
ELITE addresses this limitation through a strategy pool that evolves continuously with deployment experience.

\begin{figure*}[t]
  \centering
    \centerline{\includegraphics[width=\textwidth]{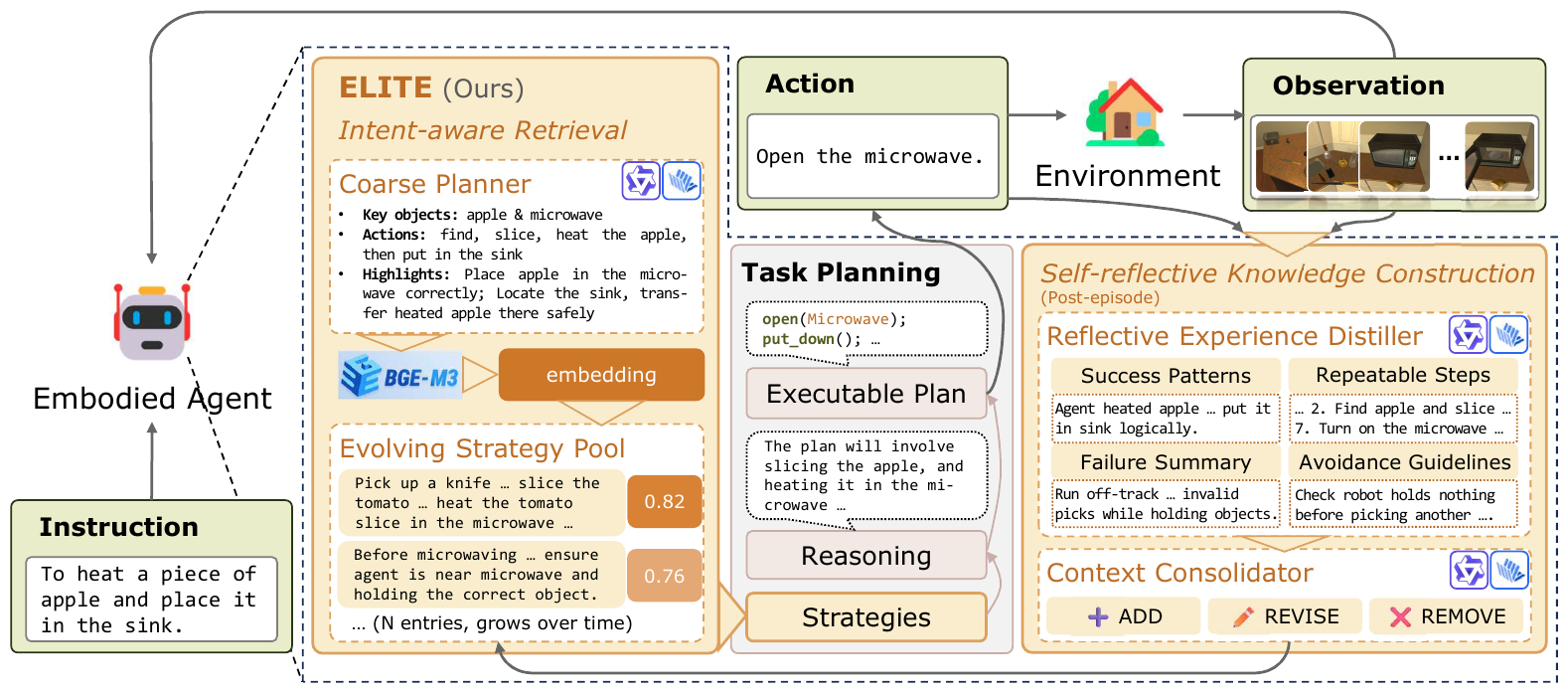}}
    \caption{
        \textbf{Overview of the ELITE framework.} The framework comprises two synergistic mechanisms: (1) \textit{Self-reflective Knowledge Construction}, which distills reusable knowledge from execution trajectories and maintains the strategy pool through structured refinement operations; and (2) \textit{Intent-aware Retrieval}, which embeds the coarse plan and retrieves procedurally similar strategies from an evolving strategy pool to augment task planning. This closed-loop architecture enables continuous improvement through physical interaction without supervision.
    }
    \label{fig:framework_0}
\end{figure*}

\paragraph{Training-based Embodied Agents.}
An alternative paradigm adapts foundation models through gradient updates on embodied data. Behavioral cloning on curated trajectories has shown promise for both low-level robotic control~\cite{kim2024openvla, zitkovich2023rt2} and high-level planning~\cite{liu2024embodied, fei2025embodied}. 
When expert demonstrations are scarce, reinforcement learning offers another path. RL4VLM~\cite{zhai2024rl4vlm} applies policy optimization to VLMs directly. ERA~\cite{chen2025era} and Reinforced Reasoner~\cite{wu2025reinforced} interleaves supervised pretraining with online RL. VAGEN~\cite{wang2025vagen} employs LLM-generated rewards for training. While effective, these approaches demand either extensive trajectory datasets or carefully shaped reward functions. 
In contrast, ELITE can improve the performance of embodied agents with only binary task outcomes.

\paragraph{Experience-Augmented Learning for Agents.}
Rather than updating parameters, a line of work stores past experience and retrieves it to inform future decisions. Reflexion~\cite{shinn2023reflexion} records verbal self-critiques after failed attempts, though these reflections remain bound to individual tasks. ExpeL~\cite{zhao2024expel} extracts transferable insights and maintains them in a searchable pool. Voyager~\cite{wang2023voyager} accumulates executable skills for open-ended exploration in Minecraft, while CLIN~\cite{majumder2024clin} distills causal rules that apply across environment variations. 
%
A common practice in these systems is that the agent retrieves experiences by textual descriptions.
%
This task description-level matching overlooks the shared structure between tasks with different descriptions but similar execution patterns. 
%
In contrast, ELITE retrieves by comparing planning traces, which captures procedural similarity that task descriptions alone may miss. 
Additionally, whereas most prior work operates in text-based settings or game environments, ELITE targets vision-based household scenarios that demand coordinated reasoning over navigation, object manipulation, and multi-step dependencies.

\section{Methods}

We present a framework enabling embodied agents to autonomously construct and leverage an evolving repository of strategic knowledge. Our approach addresses the fundamental limitation that VLMs lack procedural knowledge of physical environments by combining two interconnected components: self-reflective knowledge construction and intent-aware retrieval. In this section, we first formalize the problem setting, then describe the proposed framework in detail.

\subsection{Problem Formulation}

We formulate embodied tasks as Partially Observable Markov Decision Processes (POMDPs), represented by the tuple $(\mathcal{S}, \mathcal{A}, \Omega, \mathcal{T}, \mathcal{O}, L, \mathcal{R})$. Here, $\mathcal{S}$ denotes the environment state space, $\mathcal{A}$ is the action space comprising navigation and manipulation primitives, and $\Omega$ is the visual observation space. The transition dynamics $\mathcal{T}: \mathcal{S} \times \mathcal{A} \rightarrow \mathcal{S}$ govern state evolution, while $\mathcal{O}: \mathcal{S} \rightarrow \Omega$ maps states to visual observations $o_t \in \Omega$. The agent receives a natural language instruction $L$ specifying the goal, and the reward function $\mathcal{R}: \mathcal{S} \rightarrow \{0, 1\}$ provides a binary signal indicating task completion.

A VLM-based embodied agent maintains a history of observations and actions $h_t = (o_0, a_0, o_1, a_1, \ldots, a_{t-1}, o_t)$, and selects actions according to a policy $\pi(a_t \mid L, h_t)$ parameterized by a VLM. The episode terminates when the goal is satisfied or a maximum number of steps $T_{\max}$ is reached.

The challenge lies in bridging the gap between VLMs' static semantic knowledge and the procedural knowledge required for embodied planning. Our approach addresses this by treating the agent's context not as a fixed input, but as an evolving knowledge base that accumulates through interaction.

\subsection{ELITE Framework Overview} 
Figure~\ref{fig:framework_0} illustrates the overall architecture of ELITE.
Our method maintains an evolving strategy pool $\mathcal{P} = \{(e_i, \mathbf{z}_i)\}_{i=1}^{M}$, where each entry $e_i$ represents strategic knowledge and $\mathbf{z}_i \in \mathbb{R}^d$ is its embedding vector. The pool is initialized as empty and then grows when two components operate in a closed loop:
\begin{itemize}
    \item \textbf{Self-Reflective Knowledge Construction}: 
    The Reflective Experience Distiller $\mathcal{D}$ first extracts reusable knowledge from completed task trajectories.
    Then, Context Consolidator $\mathcal{C}$ integrates distilled knowledge into the strategy pool while maintaining coherence.
    \item \textbf{Intent-Aware Retriever}: Identifies relevant strategies for new tasks based on planning intent similarity.
\end{itemize}

\subsection{Self-Reflective Knowledge Construction via Physical Interaction}

The core idea of our approach is transforming raw execution experiences into structured, reusable strategic knowledge. This proceeds in two stages: the distiller extracts insights from individual trajectories, and the consolidator integrates these insights into a coherent knowledge base.

\paragraph{Reflective Experience Distiller.} After a task is completed, the Reflective Experience Distiller performs retrospective analysis of the trajectory $\tau = \langle(o_0, a_0), \ldots, (o_T, a_T)\rangle$ with task language instruction $L$ and outcome. 
%
Specifically, the distiller employs outcome-specific prompting strategies to extract different types of knowledge from successful versus failed executions. For successful episodes, it generates:
\begin{equation}
    r_{\text{success}} = \mathcal{D}_{\text{success}}(L, \tau; \theta_{\mathcal{D}}),
\end{equation}
capturing \textit{success patterns} that led to goal achievement, and \textit{repeatable steps} (\textit{i.e.}, concrete action sequences) that can be directly applied to similar tasks. 
For failed episodes, the distiller instead produces:
\begin{equation}
    r_{\text{failure}} = \mathcal{D}_{\text{failure}}(L, \tau; \theta_{\mathcal{D}}),
\end{equation}
yielding \textit{failure summaries} that identify root causes of execution errors, and \textit{avoidance guidelines} that specify preconditions or constraints to prevent similar failures in future tasks.

\paragraph{Context Consolidator.} The Context Consolidator then integrates these reflections into the existing strategy pool. Given current pool $\mathcal{P}^{(n-1)}$ and new reflection $r$ from the distiller, it generates delta operations:
\begin{equation}
    \Delta = \mathcal{C}(r, \mathcal{P}^{(n-1)}; \theta_{\mathcal{C}}).
\end{equation}
These operations include $\textsc{Add}(e_{\text{new}})$ for inserting novel strategies, $\textsc{Revise}(i, e'_i)$ for updating existing entries with refined insights or corrections, and $\textsc{Remove}(i)$ for deleting superseded entries. The consolidator is conditioned on the full pool contents, allowing it to determine whether new knowledge is novel, redundant, or provides useful refinements, which helps maintain pool quality as it grows with deployment experience.

\subsection{Intent-Aware Retrieval Mechanism}

As the strategy pool grows, efficient retrieval of relevant knowledge becomes critical. A naive approach based on textual similarity between task descriptions fails to capture deeper structural similarities between tasks. 
For instance, ``heat the potato and place it on the counter'' and ``warm the soup and serve it in a bowl'' share critical procedural elements but with minimal lexical overlap. They both require locating a heating appliance, activating it, and transferring the heated item to the target location.
%

To address this issue, we retrieve with planning intent rather than simple task descriptions.
%
Our key insight is that, when executing a task, an agent's reasoning process reveals its underlying intent more faithfully than the task description alone. 
%
To achieve this, we utilize chain-of-thought (CoT) embeddings. 
Specifically, before executing a task with instruction $L$ and initial observation $o_0$, a coarse planner first generates a high-level plan $p = \textsc{CoarsePlanner}(L, o_0)$ that identifies key objects, outlines action sequences, and highlights critical execution considerations. This plan is then encoded using BGE-M3~\cite{chen2024bge} to obtain a query embedding:
\begin{equation}
    \mathbf{q} = \textsc{Embed}(p).
\end{equation}
%
When the consolidator adds entry $e_i$, we also store $\mathbf{z}_i = \textsc{Embed}(p_i)$, where $p_i$ is the planning trace from the task that generated this entry. At retrieval time, we identify the top-$k$ entries by cosine similarity:
\begin{equation}
    \mathcal{M}_k = \underset{S \subset \mathcal{P}, |S|=k}{\arg\max} \sum_{(e_i, \mathbf{z}_i) \in S} \frac{\mathbf{q} \cdot \mathbf{z}_i}{\|\mathbf{q}\| \|\mathbf{z}_i\|}.
\end{equation}
The retrieved strategies $\mathcal{M}_k$ are formatted as a structured knowledge section within the planner's prompt. The planner is instructed to selectively apply these strategies based on their relevance to the current context, ensuring that marginally relevant entries do not degrade performance.

\subsection{Online Adaptation}

Our framework can operate fully online. Embodied agents can be continuously improved through interaction without ground-truth supervision. Algorithm~\ref{alg:online} presents the overall procedure. Specifically, the pool begins empty, with the agent relying solely on base VLM capabilities. This cold-start phase typically yields suboptimal performance but generates a valuable learning signal. As tasks are completed, the pool accumulates diverse knowledge spanning different task categories and execution contexts—capturing both abstract procedural strategies and concrete action sequences applicable to specific scenarios.

\begin{algorithm}[t]
\caption{Online Adaptation with Evolving Strategy Pool}
\label{alg:online}
\begin{algorithmic}[1]
\REQUIRE Task set $\mathcal{L} = \{L_1, L_2, \ldots, L_N\}$, initial pool $\mathcal{P}^{(0)} = \emptyset$, retrieval count $k$
\STATE Shuffle task set $\mathcal{L}$
\FOR{episode $n = 1, 2, \ldots, N$}
    \STATE Select task $L_n$ from $\mathcal{L}$; obtain initial observation $o_0$
    \STATE Generate coarse plan $p = \textsc{CoarsePlanner}(L_n, o_0)$
    \STATE Compute query embedding $\mathbf{q} = \textsc{Embed}(p)$
    \STATE Retrieve top-$k$ strategies $\mathcal{M}_k$ from $\mathcal{P}^{(n-1)}$ by similarity to $\mathbf{q}$
    \STATE Execute task with planner conditioned on $\mathcal{M}_k$; collect trajectory $\tau$ and outcome
    \STATE Generate reflection $r = \mathcal{D}(L_n, \tau, \text{outcome})$
    \STATE Compute delta operations $\Delta = \mathcal{C}(r, \mathcal{P}^{(n-1)})$
    \STATE Update pool: $\mathcal{P}^{(n)} = \textsc{ApplyDelta}(\mathcal{P}^{(n-1)}, \Delta)$
    \STATE Generate embeddings for new/revised entries
\ENDFOR
\end{algorithmic}
\end{algorithm}

\begin{table*}[t]
    \tablestyle{18.4pt}{1.0}
    \caption{\textbf{Results in the online unsupervised setting on EB-ALFRED and EB-Habitat.} We report the Success Rate (\%) of ELITE compared to zero-shot base VLMs and the scene-graph augmented baseline (ESCA). ELITE consistently improves over the base models without using any ground-truth trajectories. Subscripts indicate the absolute performance improvement relative to the corresponding base model.}
    \label{tab:model-performance}
    \begin{center}
        \begin{small}
            \begin{tabular*}{\textwidth}{lllllll}
                \toprule
                
                \multirow{2}{*}{\textsc{Model}} & \multicolumn{3}{c}{\textsc{EB-Alfred}} & \multicolumn{3}{c}{\textsc{EB-Habitat}} \\
                \cmidrule(lr){2-4} \cmidrule(lr){5-7}
                 & \textbf{Avg} & \textbf{Base} & \textbf{Long} & \textbf{Avg} & \textbf{Base} & \textbf{Long} \\
                
                \midrule
                

                
                
                
                \multicolumn{7}{c}{\textit{Open-Source MLLMs}} \\
                \midrule
                
                Ovis2-34B   & 29 & 34 & 24 & 41 & 68 & 14 \\
            \gr    Llama-3.2-90B-Vision-Ins & 27 & 38 & 16 & 54 & \textbf{94} & 14 \\
                gemma-3-27b-it & 34 & 42 & 26 & 45 & 68 & 22 \\
            \gr    InternVL2\_5-78B    & 40 & 38 & 42 & 54 & 80 & 28 \\
                InternVL3-78B    & 37 & 38 & 36 & 62 & 84 & \underline{40} \\
            \gr    Qwen2-VL-72B-Ins   & 35 & 40 & 30 & 44 & 70 & 18 \\
                Qwen2.5-VL-72B-Ins & \underline{52} & \underline{55} & \underline{49} & 59 & 82 & 36 \\

                \midrule

            \gr    ESCA (Qwen2.5-VL-72B-Ins) & 38 \down{14} & 46 \down{9} & 30 \down{19} & 60 \up{1} & 86 \up{4} & 34 \down{2} \\
                \textbf{Ours (Qwen2.5-VL-72B-Ins)} & \textbf{61} \up{9} & \textbf{60} \up{5} & \textbf{62} \up{13} & \underline{63} \up{4} & 86 \up{4} & \underline{40} \up{4} \\
            \gr    \textbf{Ours (InternVL3-78B)}      & 44 \up{7}       & 40 \up{2}       & 48 \up{12}        & \textbf{67}  \up{5} & \underline{90} \up{6} & \textbf{44} \up{4} \\
                                    
                \bottomrule
            \end{tabular*}
        \end{small}
    \end{center}
\end{table*}

\paragraph{Extension to Supervised Settings.} While the online setting demonstrates our framework's ability to learn without supervision, ELITE can be naturally extended to the supervised setting where ground-truth trajectories are available. 
In this setting, the Reflective Experience Distiller can perform \textit{comparative reflection}, analyzing discrepancies between the agent's trajectory and ground truth to identify specific failure causes and derive more precise corrective strategies. 

\begin{table*}[t]
    \caption{\textbf{Results in the supervised setting on EB-ALFRED.} We evaluate on both seen task categories (Base, Complex, Visual; orange) and unseen categories (Common, Spatial; green) to assess generalization. ELITE consistently improves over the base VLM across all categories and achieves the state-of-the-art performance compared with other training-based methods without requiring gradient updates. Subscripts indicate the absolute performance improvement relative to the corresponding base model.}
    \label{tab:supervised-performance}
    \begin{center}
        \begin{small}
        \begin{tabularx}{\textwidth}{lYYYYYY}
            \toprule
            
            \textbf{Model} & \textbf{Avg} & \cellcolor{myorange}\textbf{Base} & \cellcolor{myorange}\textbf{Complex} & \cellcolor{myorange}\textbf{Visual} & \cellcolor{mygreen}\textbf{Common} & \cellcolor{mygreen}\textbf{Spatial} \\
            
            \midrule
            
            \multicolumn{7}{c}{\textit{Proprietary MLLMs}} \\
            \midrule
            GPT-4o               & 56.8 & 64 & 68 & 46 & 54 & 52 \\
        \gr    Claude-3.5-Sonnet    & \underline{66.4} & \underline{72} & \underline{76} & 60 & \underline{66} & 58 \\
            Gemini-1.5-Pro       & 63.2 & 70 & 72 & 58 & 64 & 52 \\
        \gr    Gemini-2.0-flash     & 51.2 & 62 & 54 & 46 & 48 & 46 \\
            Qwen-VL-Max          & 43.2 & 44 & 44 & 42 & 48 & 38 \\

            \midrule
            \multicolumn{7}{c}{\textit{Open-Source MLLMs}} \\
            \midrule
            Ovis2-34B            & 29.6 & 34 & 38 & 28 & 30 & 18 \\
        \gr    Llama-3.2-90B-Vision & 35.2 & 38 & 44 & 28 & 34 & 32 \\
            gemma-3-27b-it       & 39.2 & 42 & 48 & 30 & 40 & 36 \\
        \gr    InternVL2\_5-78B     & 36.8 & 38 & 42 & 34 & 34 & 36 \\
            InternVL3-78B        & 39.6 & 38 & 46 & 42 & 34 & 38 \\
        \gr    Qwen2-VL-72B-Ins     & 34.4 & 40 & 40 & 30 & 30 & 32 \\
            Qwen2.5-VL-72B-Ins   & 49.8 & 55 & 55 & 50 & 42 & 47 \\
            
            \midrule
            \multicolumn{7}{c}{\textit{Training-based MLLMs}} \\
            \midrule
            RL4VLM (3B)              & 51.2 & 70 & 70 & 56 & 32 & 28 \\
        \gr    VAGEN (3B)               & 52.8 & 70 & 70 & 58 & 38 & 28 \\
            Reinforced Reasoner (7B) & 41.6 & 54 & 46 & 28 & 42 & 38 \\
        \gr    ERA (3B)                 & 65.2 & \underline{72} & 72 & \underline{62} & 54 & \textbf{66} \\
            
            \midrule
            \textbf{Ours (Qwen2.5-VL-72B-Ins)} & \quad\; \textbf{70.8} \up{21} & \quad\;\,\textbf{78} \up{23} & \quad\;\,\textbf{78} \up{23} & \quad\;\,\textbf{68} \up{18} & \quad\;\,\textbf{68} \up{26} & \quad\;\,\underline{62} \up{15} \\
            \bottomrule
        \end{tabularx}
        \end{small}
    \end{center}
\end{table*}

\section{Experiments}
\label{sec:experiments}

We evaluate ELITE on two embodied AI benchmarks to assess its ability to learn from experience and transfer knowledge across tasks. Our experiments address three key questions: (1) \textit{Can ELITE improve performance through fully online learning without supervision?} (2) \textit{How does it compare to state-of-the-art training-based methods when ground-truth trajectories are available?} and (3) \textit{Which components contribute most to its effectiveness?}

\subsection{Experimental Setup}

\paragraph{Benchmarks.} We evaluate our method on EB-ALFRED and EB-Habitat~\cite{yang2025embodiedbench}, both assessing embodied capabilities in household environments. EB-ALFRED is built on the AI2-THOR simulator~\cite{kolve2017ai2} and covers 7 household task types utilizing 8 high-level skills. It features a dynamic action space. EB-Habitat, based on Habitat 2.0~\cite{szot2021habitat}, focuses on object rearrangement tasks that involve 70 skills across five categories, including navigation and manipulation. A key distinction of EB-Habitat is its constraint on navigation targets to receptacles, necessitating multi-step search strategies. Both benchmarks provide egocentric visual observations and organize tasks into six fine-grained subsets (Base, Common Sense, Complex Instruction, Spatial Awareness, Visual Appearance, and Long Horizon) to assess distinct agent capabilities. In the online unsupervised setting, the task set is randomly shuffled, and the agent encounters each task exactly once without ground-truth supervision. Since the agent cannot retry failed tasks, performance improvements must stem from transferring knowledge across different tasks rather than from repeated attempts on the same task.

\paragraph{Baselines.} We compare against several categories of methods: (1) \textit{Open-source VLMs} including Qwen2.5-VL-72B~\cite{bai2025qwen25vl}, InternVL3-78B~\cite{zhu2025internvl3}, and others, which represent strong zero-shot baselines; (2) \textit{Proprietary VLMs} including GPT-4o, Claude-3.5-Sonnet, and Gemini-1.5-Pro in the supervised setting; (3) \textit{Training-based methods} including RL4VLM~\cite{zhai2024rl4vlm}, VAGEN~\cite{wang2025vagen}, Reinforced Reasoner~\cite{wu2025reinforced}, and ERA~\cite{chen2025era} that adapt foundation models through gradient updates; and (4) \textit{Scene graph augmentation} methods such as ESCA~\cite{huang2025esca}.

\paragraph{Implementation Details.} We implement ELITE using Qwen2.5-VL-72B-Instruct and InternVL3-78B as base VLMs for the Reflective Experience Distiller, Context Consolidator, and task planning. The coarse planner generates high-level plans with 3-6 bullet points, which are embedded using BGE-M3~\cite{chen2024bge} to obtain 1024-dimensional vectors. We retrieve $k=4$ strategies from the pool at each episode. The strategy pool is initialized to an empty set and updated after each task is completed. For the supervised setting, the distiller performs comparative reflection by analyzing discrepancies between the agent's trajectory and provided ground-truth demonstrations. All experiments use greedy decoding for the Reflective Experience Distiller, Context Consolidator, and task planning with temperature $0$ for reproducibility.

\subsection{Main Results}

\paragraph{Online Learning without Supervision.} Table~\ref{tab:model-performance} presents results in the online setting where ELITE learns purely from task outcomes without ground-truth trajectories. On EB-ALFRED, ELITE with Qwen2.5-VL-72B achieves an average success rate of 61\%, improving by 9\% over the base VLM (52\%). 
On EB-Habitat, ELITE with InternVL3-78B reaches 67\% average success (+5\%), with consistent gains across both base and long-horizon scenarios.

Notably, ELITE substantially outperforms ESCA, a recently proposed method that augments VLMs with scene graph information. While ESCA improves spatial reasoning, it cannot update its knowledge based on execution experience. On EB-ALFRED, ELITE achieves 61\% compared to ESCA's 38\%. We observe that ESCA underperforms the base VLM (52\%), possibly because the scene graph may contain some distracting information, which is a phenomenon also noted in the original ESCA paper. ELITE's experiential learning provides complementary benefits beyond structured environmental representations.

\paragraph{Supervised Setting with Ground-Truth Trajectories.} Table~\ref{tab:supervised-performance} shows results when ground-truth demonstrations are available for comparative reflection. ELITE achieves an average success rate of 70.8\% on EB-ALFRED, surpassing all open-source VLMs and proprietary models. The framework excels particularly on seen task categories (Base: 78\%, Complex: 78\%, Visual: 68\%), exceeding Claude-3.5-Sonnet with a smaller open-source base model.

Crucially, ELITE demonstrates strong generalization to unseen task categories (Common: 68\%, Spatial: 62\%), outperforming all open-source baselines by substantial margins and surpassing proprietary models. While ERA~\cite{chen2025era}, a training-based method using both SFT and RL, achieves slightly higher performance on spatial reasoning (66\%), ELITE remains competitive without requiring gradient updates or carefully shaped reward functions. This suggests that explicit knowledge distillation and retrieval can rival end-to-end training for embodied agents.


\begin{figure}[t]
  \centering
    \centerline{\includegraphics[width=\columnwidth]{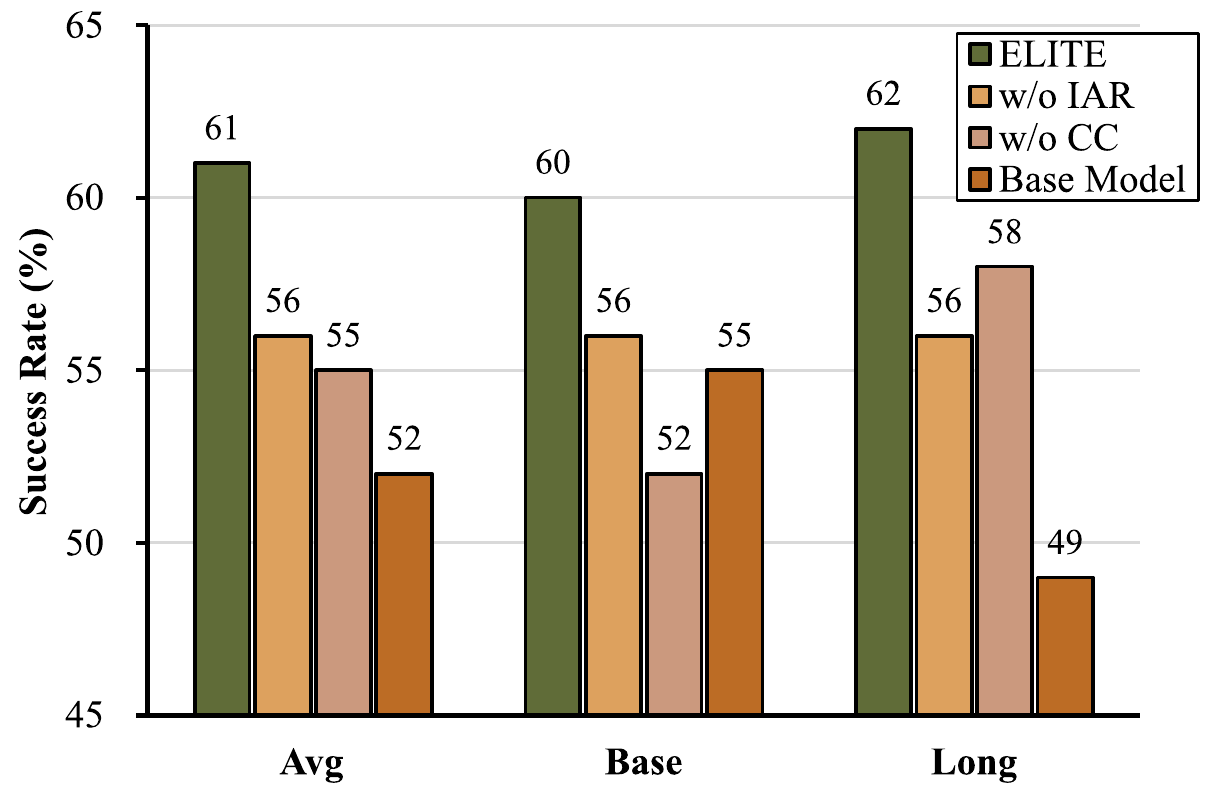}}
    \caption{
      \textbf{Ablation study on ELITE components in the online setting on EB-ALFRED.} We compare the full ELITE framework against variants without Intent-Aware Retrieval (w/o IAR), without Context Consolidation (w/o CC), and the Base Model (Qwen2.5-VL-72B). The full ELITE achieves the best results across all task categories.
    }
    \label{ablation_0}
\end{figure}

\begin{figure}[t]
  \centering
    \centerline{\includegraphics[width=0.97\columnwidth]{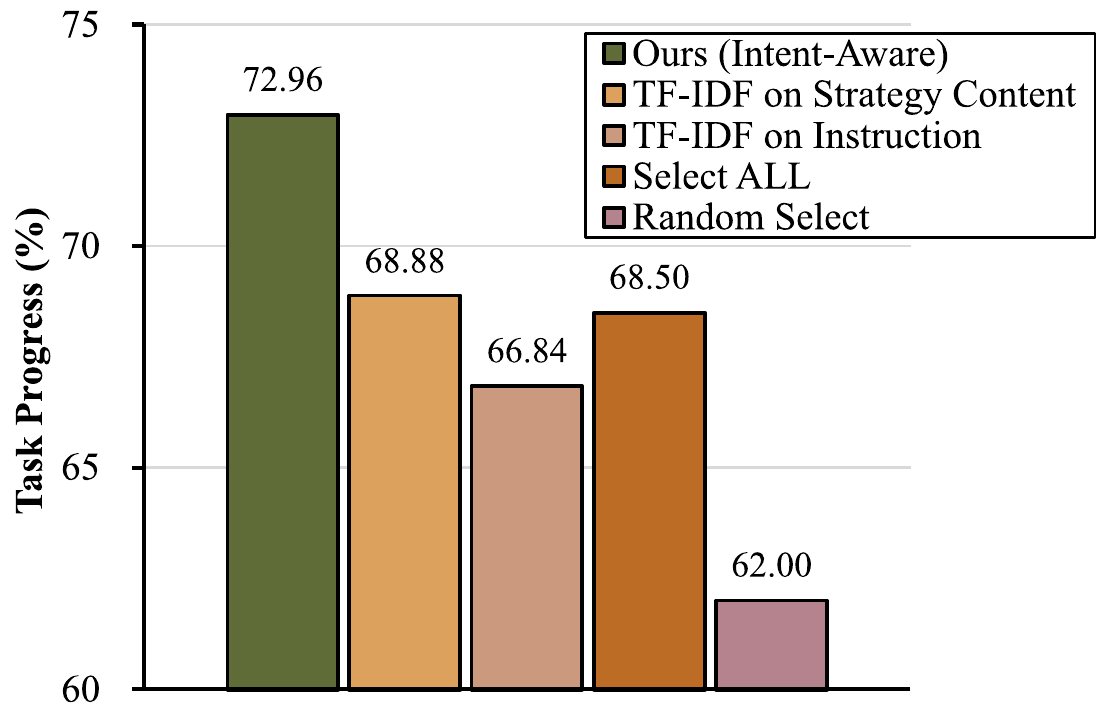}}
    \caption{
        \textbf{Ablation study on retrieval mechanisms in the online setting on EB-ALFRED long-horizon tasks.} Task progress denotes the average completion percentage across all long-horizon tasks. Our intent-aware (CoT) retrieval outperforms TF-IDF-based alternatives (strategy content and instruction similarity), as well as using all strategies in the strategy pool and random selection.
    }
    \label{ret_abl_0}
\end{figure}

\subsection{Ablation Studies}

\paragraph{Component Analysis.} Figure~\ref{ablation_0} analyzes the contribution of each component by removing: (1) Intent-Aware Retrieval (w/o IAR), which uses the full set of strategies; and (2) Context Consolidation (w/o CC), which adds reflections directly without refinement. Both components prove essential. Specifically, without intent-aware retrieval, performance drops to 56\% average on EB-ALFRED. This suggests that simply using all available strategies may introduce irrelevant noise, potentially distracting the planner. Similarly, removing consolidation results in a performance decline to 55\%. We hypothesize that raw reflections often contain unrefined reasoning details; without the consolidation process to distill this information into concise insights, the planner may struggle to utilize the context effectively. 

\begin{figure}[t]
  \centering
    \centerline{\includegraphics[width=0.95\columnwidth]{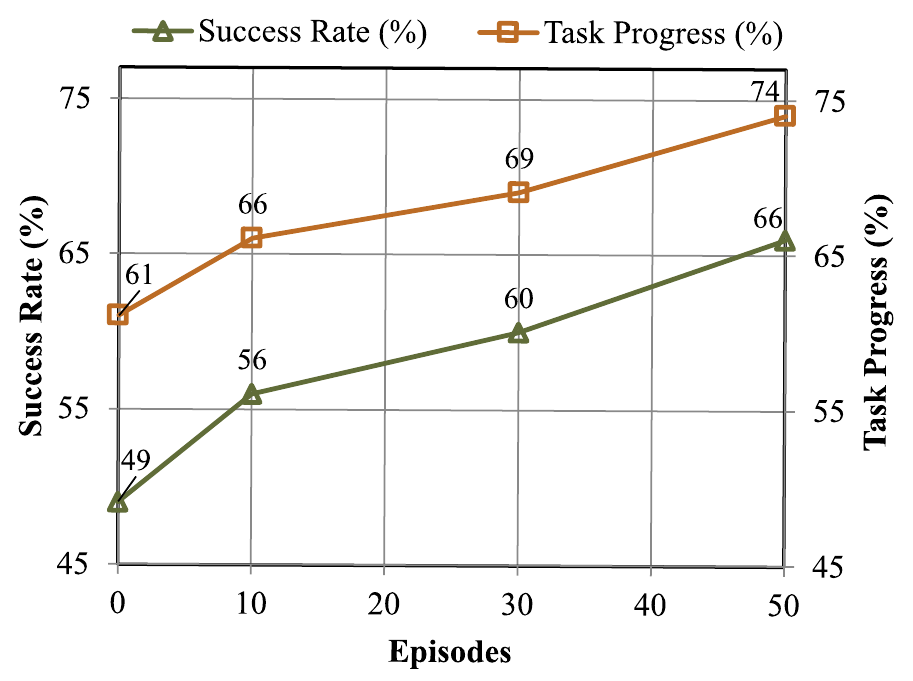}}
    \caption{
        \textbf{Illustration of learning dynamics of ELITE in the online setting on EB-ALFRED long-horizon tasks.} The $x$-axis indicates the number of tasks processed for online learning, while the $y$-axis shows the average success rate and task progress across all long-horizon tasks. Both metrics improve consistently as the strategy pool accumulates experience, demonstrating continuous self-improvement through deployment experience.
    }
    \label{curve_0}
\end{figure}

\begin{figure*}[t]
  \centering
    \centerline{\includegraphics[width=\textwidth]{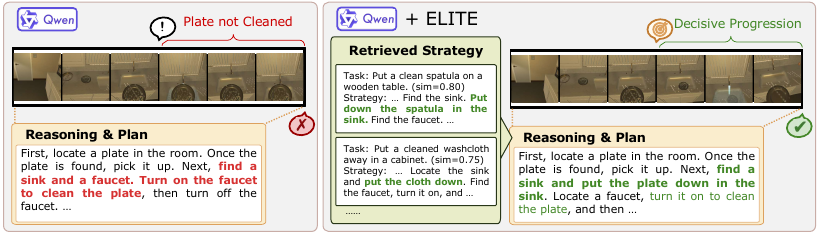}}
    \caption{
        \textbf{Qualitative comparison between the base Qwen2.5-VL model and ELITE on the example task: ``Put a clean plate on the counter.''} \textit{Left:} The base model generates a flawed plan that attempts to clean the plate without first placing it in the sink, resulting in task failure. \textit{Middle:} ELITE retrieves procedurally similar strategies from past experiences (sim=0.80 and 0.75) that demonstrate the correct pattern of putting objects in the sink before cleaning. \textit{Right:} Augmented with retrieved strategies, ELITE produces a corrected plan that properly places the plate in the sink before turning on the faucet, leading to successful task completion.
    }
    \label{traj_0}
\end{figure*}

\paragraph{Retrieval Mechanism Comparison.} Figure~\ref{ret_abl_0} compares our intent-aware retrieval against alternative mechanisms in the online setting on EB-ALFRED long-horizon tasks. We report task progress, defined as the average completion percentage across all long-horizon tasks. Our CoT-based approach achieves 72.96\% task progress, substantially outperforming retrieval based on TF-IDF, a standard sparse lexical matching method that retrieves strategies by either strategy content (68.88\%) or instruction similarity (66.84\%). Notably, using all available strategies without selective retrieval yields only 68.50\%, confirming that an indiscriminate strategy introduces noise and degrades planning quality. Random retrieval performs worst at 62.00\%, establishing a lower bound. These results validate our hypothesis that planning traces capture procedural similarity more effectively than sparse lexical matching, enabling transfer between tasks with different descriptions but shared execution patterns.

\paragraph{Learning Dynamics.} Figure~\ref{curve_0} illustrates how ELITE's performance evolves as it accumulates experience in the online setting on EB-ALFRED long-horizon tasks. The $x$-axis denotes the number of tasks processed for online learning, while the $y$-axis reports the average performance across all long-horizon tasks, including seen and unseen tasks. Specifically, with an empty strategy pool at episode 0, the agent achieves a 49\% success rate and 61\% task progress, reflecting the base VLM's capabilities. As the agent encounters more tasks and populates its strategy pool, both metrics improve consistently. After 10 episodes, the success rate rises to 56\% with 66\% task progress; after 30 episodes, these reach 60\% and 69\%, respectively. By episode 50, ELITE achieves a 66\% success rate and 74\% task progress, representing gains of 17\% and 13\% over the cold-start baseline. The monotonic improvement demonstrates that our framework enables continuous self-improvement through deployment experience.

\subsection{Qualitative Analysis}

Figure~\ref{traj_0} illustrates how ELITE guides task execution through retrieved strategies. Given the instruction ``Put a clean plate on the counter,'' the base Qwen2.5-VL model generates a flawed plan that attempts to clean the plate before placing it in the sink. After experiencing similar tasks and accumulating strategies like ``Find the sink. Put down the spatula in the sink'' and ``Locate the sink and put the cloth down,'' ELITE retrieves these procedural, similar experiences. The augmented planner then generates a corrected plan: ``find a sink and put the plate down in the sink,'' followed by proper cleaning steps. This demonstrates how our framework captures procedural dependencies (\textit{e.g.,} objects must be placed in appropriate locations for manipulation) that foundation models alone struggle to internalize.

The retrieved strategies exhibit high semantic similarity to the current task (sim=0.80 and 0.75) despite different task descriptions, validating our intent-aware retrieval mechanism. By comparing planning traces rather than instruction text, ELITE successfully identifies that ``put a clean spatula on a wooden table'' and ``put a cleaned washcloth away in a cabinet'' share the critical pattern of sink-based cleaning, enabling effective knowledge transfer.

\section{Conclusion}

This paper presents ELITE, a framework that enables embodied agents to autonomously accumulate and leverage experiential knowledge. By extracting reusable strategies from execution trajectories and retrieving them based on procedural intent rather than task descriptions, ELITE bridges the gap between foundation models' semantic understanding and the procedural knowledge required for reliable physical interaction. Experiments demonstrate that the proposed framework achieves substantial improvements over base VLMs in fully online settings (+9\% on EB-ALFRED, +5\% on EB-Habitat) without any supervision, and state-of-the-art performance compared with training-based methods when ground-truth trajectories are available.

\noindent \textbf{Limitations.} First, our intent-aware retrieval relies on the quality of the coarse planner's reasoning traces, meaning errors in initial planning can propagate to retrieval. Second, the single global strategy pool may benefit from hierarchical or task-specific organization for improved scalability. Third, extending beyond household tasks with discrete actions to continuous control remains an open challenge.

Despite these limitations, ELITE demonstrates that embodied agents can be improved through self-directed experiential learning. As foundation models continue to advance, frameworks that enable agents to ground their capabilities through physical interaction will become increasingly important for deploying reliable autonomous systems in the real world.

\section*{Impact Statement}

This work advances embodied AI by enabling agents to learn from experience and improve autonomously. While this has potential benefits for assistive robotics, manufacturing, and household automation, several considerations warrant attention.

Self-improving agents could propagate incorrect strategies if learned knowledge is flawed, requiring rigorous validation for safety-critical deployments. Embodied agents in private spaces must carefully handle data collection to ensure learned strategies do not encode sensitive information. More capable agents could affect employment in physical task domains, though this reflects broader automation trends, while also offering benefits, including improved accessibility for individuals with disabilities and reduced human exposure to dangerous tasks.

We believe this work represents meaningful progress toward adaptive embodied agents and encourage continued attention to safety, transparency, and ethical deployment as the field advances.


\bibliography{references}
\bibliographystyle{icml2026}

\newpage
\appendix
\onecolumn

\end{document}